\documentclass{article}
\pdfoutput=1
\PassOptionsToPackage{colorlinks,linkcolor={red!80!black},citecolor={blue!70!black},urlcolor={blue!80!black},hypertexnames=false}{hyperref}

\usepackage{amsmath,amssymb}
\usepackage{color}
\usepackage{xcolor}
\usepackage{fullpage}
\usepackage{graphicx}

\usepackage[style=authoryear]{biblatex}
\usepackage{hyperref}
\usepackage[capitalize]{cleveref}

\DeclareMathOperator{\E}{\mathbb E}
\newcommand{\h}{\mathcal H}
\DeclareMathOperator{\N}{\mathcal N}
\DeclareMathOperator{\Tr}{\mathrm{Tr}}
\newcommand{\ud}{\mathrm d}

\makeatletter
\newcommand\given{%
 \@ifstar
  {\mathrel{}\middle|\mathrel{}}
  {\mid}%
}
\DeclareRobustCommand{\abs}{\@ifstar\@abs\@@abs}
\newcommand{\@abs}[1]{\left\lvert #1 \right\rvert}
\newcommand{\@@abs}[1]{\lvert #1 \rvert}

\DeclareRobustCommand{\norm}{\@ifstar\@norm\@@norm}
\newcommand{\@norm}[1]{\left\lVert #1 \right\rVert}
\newcommand{\@@norm}[1]{\lVert #1 \rVert}

\DeclareRobustCommand{\ceil}{\@ifstar\@ceil\@@ceil}
\newcommand{\@ceil}[1]{\left\lceil #1 \right\rceil}
\newcommand{\@@ceil}[1]{\lceil #1 \rceil}

\DeclareRobustCommand{\floor}{\@ifstar\@floor\@@floor}
\newcommand{\@floor}[1]{\left\lfloor #1 \right\rfloor}
\newcommand{\@@floor}[1]{\lfloor #1 \rfloor}
\makeatother

\title{Fixing an error in \texorpdfstring{\textcite{caponnetto}}{Caponnetto and de Vito (2007)}}
\author{Danica J.\ Sutherland\\{\normalsize Gatsby unit, University College London}\\{\normalsize \texttt{djs@djsutherland.ml}}}
\date{February 2017}

\begin{document}
\maketitle

\begin{abstract}
The seminal paper of \textcite{caponnetto} provides minimax-optimal rates for kernel ridge regression in a very general setting.
Its proof, however, contains an error in its bound on the effective dimensionality.
In this note, we explain the mistake,
provide a correct bound,
and show that the main theorem remains true.
\end{abstract}

The mistake lies in Proposition 3's bound on the effective dimensionality $\mathcal N(\lambda)$,
particularly its dependence on the parameters of the family of distributions $b$ and $\beta$.
We discuss the mistake and provide a correct bound in \cref{sec:n-lambda}.
Its dependence on the regularization parameter $\lambda$, however, was correct, so the proof of Theorem 1 carries through with the exact same strategy.
The proof was written in such a way, though, that it is not immediately obvious that it still holds for the corrected bound;
we thus provide a more detailed explication of the proof, showing it is still valid.

This note will make little sense without a copy of the original paper at hand.
Numbered theorems and equation references always refer to those of \textcite{caponnetto};
equations in this document are labeled alphabetically.

\paragraph{A trivial correction}
First, we note a tiny mistake:
Theorem 4 needs $C_\eta = 96 \log^2 \frac{6}{\eta}$, rather than $32 \log^2 \frac{6}{\eta}$, because the last line of its proof dropped the constant 3 in front of $S_1(\lambda, \mathbf z)$ and $S_2(\lambda, \mathbf z)$ in (36).

\section{Bound on the effective dimensionality} \label{sec:n-lambda}

Part of Proposition 3 is the claim that
for $p \in \mathcal P(b, c)$, with $c \in [1, 2]$ and $b \in (1, \infty)$,
\begin{equation}
    \N(\lambda) \le \frac{\beta b}{b - 1} \lambda^{-\frac1b}
    \label{eq:claimed-bound}
.\end{equation}
The argument starts like this:
\begin{align}
       \mathcal N(\lambda)
  &=   \Tr\left[ (T + \lambda I)^{-1} T \right]      \notag
\\&=   \sum_{n=1}^\infty \frac{t_n}{t_n + \lambda}      \notag
\\&\le \sum_{n=1}^\infty \frac{\beta / n^b}{\beta / n^b + \lambda}
   =   \sum_{n=1}^\infty \frac{\beta}{\beta + \lambda n^b} \label{eq:sum-bound}
\\&\le \int_0^\infty\! \frac{\beta}{\beta + \lambda x^b} \,\ud x      \notag
\\&=   \lambda^{-\tfrac1b} \int_0^\infty\! \frac{\beta}{\beta + \tau^b} \,\ud \tau
\label{eq:bound-int}
,\end{align}
with \eqref{eq:sum-bound} following from Definition 1 (iii),
the next line's upper bound by an integral true since $x \mapsto \frac{\beta}{\beta + \lambda x^b}$ is decreasing,
and then doing a change of variables to $\tau^b = \lambda x^b$.

But then the paper claims without further justification that
\begin{equation}
    \int_0^\infty\! \frac{1}{\beta + \tau^b} \,\ud\tau \le \frac{b}{b-1}
    \label{eq:wrong}
.\end{equation}
In fact, \eqref{eq:wrong} is incorrect:
as $\beta \to 0$, the integral approaches the divergent integral $\int_0^\infty \tau^{-b} \ud\tau$,
but $\frac{b}{b-1}$ clearly does not depend on $\beta$.
We can instead compute the true value of the integral:
\begin{align}
    \int_0^\infty \frac{1}{\beta + \tau^b} \ud\tau
  &= \int_0^\infty \frac{1 / \beta}{1 + \left( \tau \beta^{-\frac1b} \right)^b} \ud\tau  \notag
\\&= \beta^{-1} \int_0^\infty \frac{1}{1 + u^b} \beta^\frac1b \ud u  \notag
\\&= \beta^\frac{1-b}{b} \frac{\pi / b}{\sin(\pi / b)}
    \label{eq:int-actual}
.\end{align}
Using \eqref{eq:int-actual} in \eqref{eq:bound-int},
we get a correct lower bound:
\begin{equation}
    \mathcal N(\lambda)
    \le \beta^\frac1b \frac{\pi / b}{\sin(\pi / b)} \lambda^{-\frac1b}
    \label{eq:n-bound}
.\end{equation}
\eqref{eq:n-bound} has the same dependence on $\lambda$ as \eqref{eq:claimed-bound},
but the dependence on $\beta$ and $b$ differs.

To demonstrate, we now plot the sum \eqref{eq:sum-bound} (green, middle),
the correct upper bound \eqref{eq:n-bound} (blue, top),
and the purported upper bound \eqref{eq:claimed-bound} obtained via \eqref{eq:wrong} (orange, bottom)
for $\beta = 0.1, \lambda = 10^{-3}$.
\begin{center}
\includegraphics[width=\textwidth]{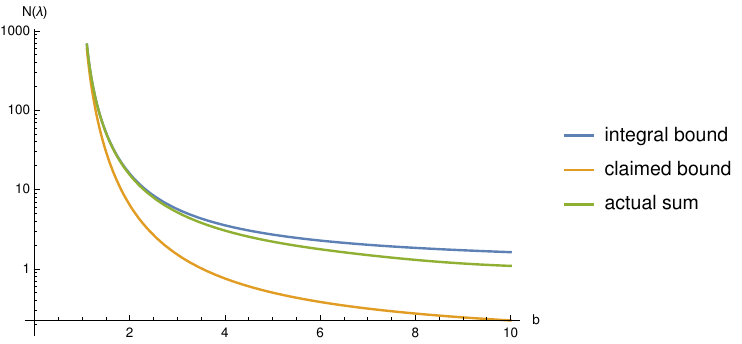}
\end{center}

\section{Consequences in Theorem 1}
Though it isn't obvious at first, the proof of Theorem 1 depends on the $\mathcal N(\lambda)$ bound only in its rate on $\lambda$, which was indeed correct: thus the proof of Theorem 1 remains valid.
We now restate the relevant parts of the proof of Theorem 1 in a way that makes this lack of dependence more explicit.

Theorem 4 gives us that for any any $\eta \in (0, 1)$, with probability greater than $1 - \eta$ we have
\begin{align*}
       \mathcal E[f_{\mathbf z}^\lambda] - \mathcal E[f_{\mathcal H}]
  &\le C_\eta\left(
          \mathcal A(\lambda)
        + \frac{\kappa^2}{\ell^2 \lambda} \mathcal B(\lambda)
        + \frac{\kappa}{\ell \lambda} \mathcal A(\lambda)
        + \frac{\kappa M^2}{\ell^2 \lambda}
        + \frac{\Sigma^2}{\ell} \mathcal N(\lambda)
       \right)
,\end{align*}
provided that
\[
    \ell \ge \frac{2 C_\eta \kappa}{\lambda} \mathcal N(\lambda)
    \qquad\text{and}\qquad
    \lambda \le \norm{T}_{\mathcal L(\h)}
\label{eq:conditions}
.\]

Define $Q$ as
\[
    Q = \begin{cases}
    \beta^\frac1b \frac{\pi / b}{\sin(\pi / b)} & b < \infty \\
    \beta & b = \infty
    \end{cases}
,\]
so that, from Proposition 3 and \eqref{eq:n-bound},
\[
    \mathcal A(\lambda) \le \lambda^c \norm{T^\frac{1-c}{2} f_\h}_\h^2
    \qquad\qquad
    \mathcal B(\lambda) \le \lambda^{c-1} \norm{T^\frac{1-c}{2} f_\h}_\h^2
    \qquad\qquad
    \mathcal N(\lambda) \le Q \lambda^{-\frac1b}
.\]
Plugging in these rates and $\norm{T^{\frac{1-c}{2}} f_{\h}}_{\h}^2 \le R$ from Definition 1 (ii), we have that
\begin{gather}
       \mathcal E[f_{\mathbf z}^\lambda] - \mathcal E[f_{\mathcal H}]
  \le C_\eta\left(
          R \lambda^c
        + \kappa^2 R \frac{\lambda^{c-2}}{\ell^2}
        + \kappa R   \frac{\lambda^{c-1}}{\ell}
        + \kappa M^2 \frac{\lambda^{-1}}{\ell^2}
        + \Sigma^2 Q \frac{\lambda^{-\frac1b}}{\ell}
       \right)
\label{eq:risk-bound-R}
\\
  \text{as long as}\qquad
    \ell \ge 2 C_\eta \kappa Q \lambda^{-\frac{b+1}{b}}
\label{eq:ell-requirement}
\\
    \qquad\text{and}\qquad
    \lambda \le \norm{T}_{\mathcal L(\h)}
    \notag
.\end{gather}

Note that for $b = \infty$,
nothing has changed from the paper.
Thus the proofs (which were not written explicitly in the paper) remain the same.
We thus assume $b < \infty$.

\subsection{\texorpdfstring{$c > 1$}{c > 1}}
When $c > 1$, let
\[
    \ell_\eta \ge \left( 2 C_\eta \kappa Q \right)^\frac{bc+1}{b(c-1)}
,\]
so that for any $\ell \ge \ell_\eta$ we have
\[
    \ell^\frac{b(c-1)}{bc + 1} \ge 2 C_\eta \kappa Q
.\]
Define $\lambda_\ell = \ell^{-\frac{b}{bc+1}}$.
Then
\begin{align*}
       2 C_\eta \kappa Q \lambda_\ell^{-\frac{b+1}{b}}
  &=   2 C_\eta \kappa Q \left( \ell^{-\frac{b}{bc+1}} \right)^{-\frac{b+1}{b}}
\\&=   2 C_\eta \kappa Q \ell^{\frac{b + 1}{bc+1}}
\\&=   2 C_\eta \kappa Q \ell^{- \frac{b(c-1)}{bc+1}} \ell^{\frac{b + 1}{bc+1} + \frac{b(c-1)}{bc+1}}
\\&\le \ell^{\frac{b + 1}{bc+1} + \frac{bc-b}{bc+1}}
   =   \ell
,\end{align*}
so that \eqref{eq:ell-requirement} holds for $\lambda = \lambda_\ell$,
and therefore \eqref{eq:risk-bound-R} holds with probability at least $1-\eta$
as long as $\lambda_\ell \le \norm{T}_{\mathcal L(\h)}$.
By Definition 1 (iii), the latter is at least $\alpha$;
thus this condition is met as long as
\begin{gather*}
    \lambda_\ell = \ell^{-\frac{b}{bc+1}} \le \alpha
    \qquad\text{i.e.}\;\;
    \ell \ge \alpha^{-\frac{bc+1}{b}}
.\end{gather*}
We thus don't have to worry about it in the asymptotics.
Plugging $\lambda_\ell$ into \eqref{eq:risk-bound-R}, we get
\begin{align*}
       \mathcal E[f_{\mathbf z}^\lambda] - \mathcal E[f_{\mathcal H}]
  &\le C_\eta\left(
          R \lambda_\ell^c
        + \kappa^2 R \frac{\lambda_\ell^{c-2}}{\ell^2}
        + \kappa R   \frac{\lambda_\ell^{c-1}}{\ell}
        + \kappa M^2 \frac{\lambda_\ell^{-1}}{\ell^2}
        + \Sigma^2 Q \frac{\lambda_\ell^{-\frac1b}}{\ell}
       \right)
\\&= C_\eta\left(
          R \ell^{-\frac{bc}{bc + 1}}
        + \kappa^2 R \ell^{-\frac{b(c-2)}{bc+1} - 2}
        + \kappa R \ell^{-\frac{b(c-1)}{bc+1} - 1}
        + \kappa M^2 \ell^{\frac{b}{bc+1} - 2}
        + \Sigma^2 Q \ell^{\frac{b}{b(bc+1)} - 1}
       \right)
\\&= C_\eta\left(
          R \ell^{-\frac{bc}{bc + 1}}
        + \kappa^2 R \ell^{-\frac{3 b c - 2b + 2}{bc+1}}
        + \kappa R \ell^{-\frac{2 b c - b + 1}{bc+1}}
        + \kappa M^2 \ell^{\frac{b - 2 b c - 2}{bc+1}}
        + \Sigma^2 Q \ell^{-\frac{bc}{bc+1}}
       \right)
.\end{align*}
Note that
\begin{gather*}
3 b c - 2 b + 2 = bc + 2 b (c-1) + 2 > b c \\
2 b c - b + 1 = bc + b (c-1) + 1 > b c \\
2 b c - b + 2 = bc + b(c-1) + 2 > b c
.\end{gather*}
Thus for large $\ell$ the $\ell^{-\frac{bc}{bc + 1}}$ terms dominate,
and so we have that
\[
       \mathcal E[f_{\mathbf z}^\lambda] - \mathcal E[f_{\mathcal H}]
       \le 2 C_\eta D \ell^{-\frac{bc}{bc + 1}}
       \qquad \forall \ell \ge \ell_\eta
,\]
where $D$ is some complex function of $R$, $\kappa$, $M$, $\Sigma$, $\beta$, $b$, and $c$.
Letting $\tau = 2 C_\eta D = 192 D \log^2 \frac6\eta$ and solving for $\eta$, we get
$\eta_\tau = 6 e^{-\sqrt{\frac\tau{192 D}}}$.
Thus
\[
    \Pr_{\mathbf z \sim \rho^\ell}\left[
        \mathcal E[f_{\mathbf z}^{\lambda_\ell}] - \mathcal E[f_\h]
        > \tau \ell^{-\frac{bc}{bc+1}}
    \right] \le \eta_\tau
    \qquad \forall \ell \le \ell_{\eta_\tau}
,\]
and so
\[
    \limsup_{\ell\to\infty} \sup_{\rho \in \mathcal P(b, c)}
    \Pr_{\mathbf z \sim \rho^\ell}\left[
        \mathcal E[f_{\mathbf z}^{\lambda_\ell}] - \mathcal E[f_\h]
        > \tau \ell^{-\frac{bc}{bc+1}}
    \right] \le \eta_\tau
,\]
and since $\lim_{\tau\to\infty} \eta_\tau = \lim_{\tau\to\infty} 6 e^{-\sqrt{\frac\tau{192 D}}} = 0$,
we have as desired that
\[
    \lim_{\tau\to\infty}
    \limsup_{\ell\to\infty} \sup_{\rho \in \mathcal P(b, c)}
    \Pr_{\mathbf z \sim \rho^\ell}\left[
        \mathcal E[f_{\mathbf z}^{\lambda_\ell}] - \mathcal E[f_\h]
        > \tau \ell^{-\frac{bc}{bc+1}}
    \right] = 0
.\]

\subsubsection{\texorpdfstring{$c = 1$}{c = 1}}
Here we define $\lambda_\ell = \left( \frac{\log \ell}{\ell} \right)^\frac{b}{b+1}$,
so that the $\ell$ requirement of \eqref{eq:ell-requirement} is
\[
    \ell \ge 2 C_\eta \kappa Q \frac{\ell}{\log \ell}
,\]
that is,
\[
    \ell \ge \exp\left( 2 C_\eta \kappa Q \right)
,\]
so choosing $\ell_\eta = \exp\left( 2 C_\eta \kappa Q \right)$ suffices.

As in the $c > 1$ case, plugging $\lambda_\ell$ into \eqref{eq:risk-bound-R} obtains that as long as $\ell \ge \ell_\eta$ (and $\lambda_\ell \le \alpha$),
\begin{align*}
       \mathcal E[f_{\mathbf z}^\lambda] - \mathcal E[f_{\mathcal H}]
  &\le C_\eta\left(
          R \lambda_\ell
        + \kappa^2 R \frac{\lambda_\ell^{-1}}{\ell^2}
        + \kappa R   \frac{1}{\ell}
        + \kappa M^2 \frac{\lambda_\ell^{-1}}{\ell^2}
        + \Sigma^2 Q \frac{\lambda_\ell^{-\frac1b}}{\ell}
       \right)
\\&=   C_\eta\left(
          R \left( \frac{\log \ell}{\ell} \right)^\frac{b}{b+1}
        + \kappa^2 R (\log \ell)^\frac{-b}{b+1} \ell^{\frac{b}{b+1} - 2}
        + \kappa R \ell^{-1}
        + \kappa M^2 (\log\ell)^{-\frac{b}{b+1}} \ell^{\frac{b}{b+1} - 2}
        + \Sigma^2 Q (\log\ell)^{-\frac{1}{b+1}} \ell^{\frac{1}{b+1} - 1}
       \right)
\\&=   C_\eta\left(
          R \left( \frac{\log \ell}{\ell} \right)^\frac{b}{b+1}
        + \kappa^2 R (\log \ell)^\frac{-b}{b+1} \ell^{-\frac{b + 2}{b+1}}
        + \kappa R \ell^{-1}
        + \kappa M^2 (\log\ell)^{-\frac{b}{b+1}} \ell^{-\frac{b + 2}{b+1}}
        + \Sigma^2 Q (\log\ell)^{-\frac{1}{b+1}} \ell^{-\frac{b}{b+1}}
       \right)
.\end{align*}
The $\ell^{-\frac{b}{b+1}}$ terms dominate,
and so we can find a value $D'$ such that
\[
       \mathcal E[f_{\mathbf z}^\lambda] - \mathcal E[f_{\mathcal H}]
       \le 2 C_\eta D' \ell^{-\frac{b}{b + 1}}
       \qquad \forall \ell \ge \ell_\eta
,\]
and the result follows by the same reasoning.

\printbibliography

\end{document}